\documentclass[11pt]{article}
\usepackage{graphicx}
\usepackage{coling2020}
\usepackage{times}
\usepackage{url}
\usepackage[dvipsnames, table]{xcolor}
\usepackage{makecell}
\usepackage{enumitem}
\usepackage{adjustbox}
\usepackage{comment}

\usepackage[T1]{fontenc}
\usepackage{xcolor}

\usepackage{multirow}
\usepackage{latexsym}
\usepackage{booktabs}
\usepackage{courier}
\usepackage{array}
\usepackage{colortbl}

\usepackage{amsmath}
\usepackage{subfig}
\usepackage{placeins}
\usepackage{tikz}
\usepackage{caption}
\usepackage{longtable}
\usepackage{subfig}
\usepackage{hyperref}
\usepackage{amssymb}

\newcommand{\ra}[1]{\renewcommand{\arraystretch}{#1}}

\newcommand{\tikzcircle}[2][red,fill=red]{\tikz[baseline=-0.5ex]\draw[#1,radius=#2] (0,0) circle ;}%
\setlist[itemize]{leftmargin=*}
\makeatletter
\newcommand{\printfnsymbol}[1]{%
  \textsuperscript{\@fnsymbol{#1}}%
}
\makeatother

\title{A Survey on Explainability in Machine Reading Comprehension}

\author{Mokanarangan Thayaparan\thanks{\ \ : \ \texttt{equal contribution}} ,  Marco Valentino\printfnsymbol{1},
   Andr\'e Freitas \\
  Department of Computer Science\\
  University of Manchester \\
  \texttt{\{mokanarangan.thayaparan,
  marco.valentino,
  andre.freitas\}}\\ \texttt{@manchester.ac.uk}
  \\}

\date{}

\begin{document}
\maketitle
\begin{abstract}
This paper presents a systematic review of benchmarks and approaches for \textit{explainability} in Machine Reading Comprehension (MRC). We present how the representation and inference challenges evolved and the steps which were taken to tackle these challenges. We also present the evaluation methodologies to assess the performance of explainable systems. In addition, we identify persisting open research questions and highlight critical directions for future work.
\end{abstract}

\section{Introduction}
\label{sec:intro}
\emph{Machine Reading Comprehension (MRC)} has the long-standing goal of developing machines that can reason with natural language. A typical reading comprehension task consists in answering questions about the background knowledge expressed in a textual corpus. Recent years have seen an explosion of models and architectures due to the release of large-scale benchmarks, ranging from open-domain \cite{rajpurkar2016squad,yang2018hotpotqa} to commonsense \cite{talmor2019commonsenseqa,huang2019cosmos} and scientific \cite{khot2020qasc,clark2018think} reading comprehension tasks. Research in MRC is gradually evolving in the direction of abstractive inference capabilities, going beyond what is explicitly stated in the text \cite{baral2020natural}.
As the need to evaluate abstractive reasoning becomes predominant, a crucial requirement emerging in recent years is \emph{explainability} \cite{miller2019explanation}, intended as the ability of a model to expose the underlying mechanisms adopted to arrive at the final answers. Explainability has the potential to tackle some of the current issues in the field: 
    \paragraph{Evaluation:} Traditionally, MRC models have been evaluated on end-to-end prediction tasks. In other words, the capability of achieving high accuracy on specific datasets has been considered a proxy for evaluating a desired set of reasoning skills. However, recent work have demonstrated that this is not necessarily true for models based on deep learning, which are particularly capable of exploiting biases in the data \cite{mccoy2019right,gururangan2018annotation}. Research in explainability can provide novel evaluation frameworks to investigate and analyse the internal reasoning mechanisms \cite{inoue2020r4c,dua-etal-2020-benefits,ross2017right};
    \paragraph{Generalisation:} Despite remarkable performance achieved in specific MRC tasks, machines based on deep learning still suffer from overfitting and lack of generalisation. By focusing on explicit reasoning methods, research in explainability can lead to the development of novel models able to perform compositional generalisation \cite{andreas2016learning,Gupta2020Neural} and discover abstract inference patterns in the data \cite{khot2020qasc,rajani2019explain}, favouring few-shot learning and cross-domain transportability \cite{camburu2018snli};
    \paragraph{Interpretability:} A system capable of delivering explanations is generally more interpretable, meeting some of the requirements for real world applications, such as user trust, confidence and acceptance \cite{biran2017explanation}.

Despite the potential impact of explainability in MRC, little has been done to provide a unifying and organised view of the field. This paper aims at systematically categorising explanation-supporting benchmarks and models. To this end, we review the work published in the main AI and NLP conferences from 2015 onwards which actively contributed to explainability in MRC, referring also to preprint versions when necessary. The survey is organised as follows: (a) Section 2 frames the scope of the survey, stating a definition of explainability in MRC; (b) Section 3 reviews the main benchmarks proposed in recent years for the explicit evaluation and development of explainable MRC models; (c) Section 4 provides a detailed classification of the main architectural patterns and approaches proposed for explanation generation; (d) Section 5 describes quantitative and qualitative metrics for the evaluation of explainability, highlighting some of the issues connected with the development of explanation-supporting benchmarks. 

\section{Explainability in Machine Reading Comprehension}

\label{sec:explanation_as_nli}

\begin{figure}[t]
\centering
\includegraphics[width=0.95\textwidth]{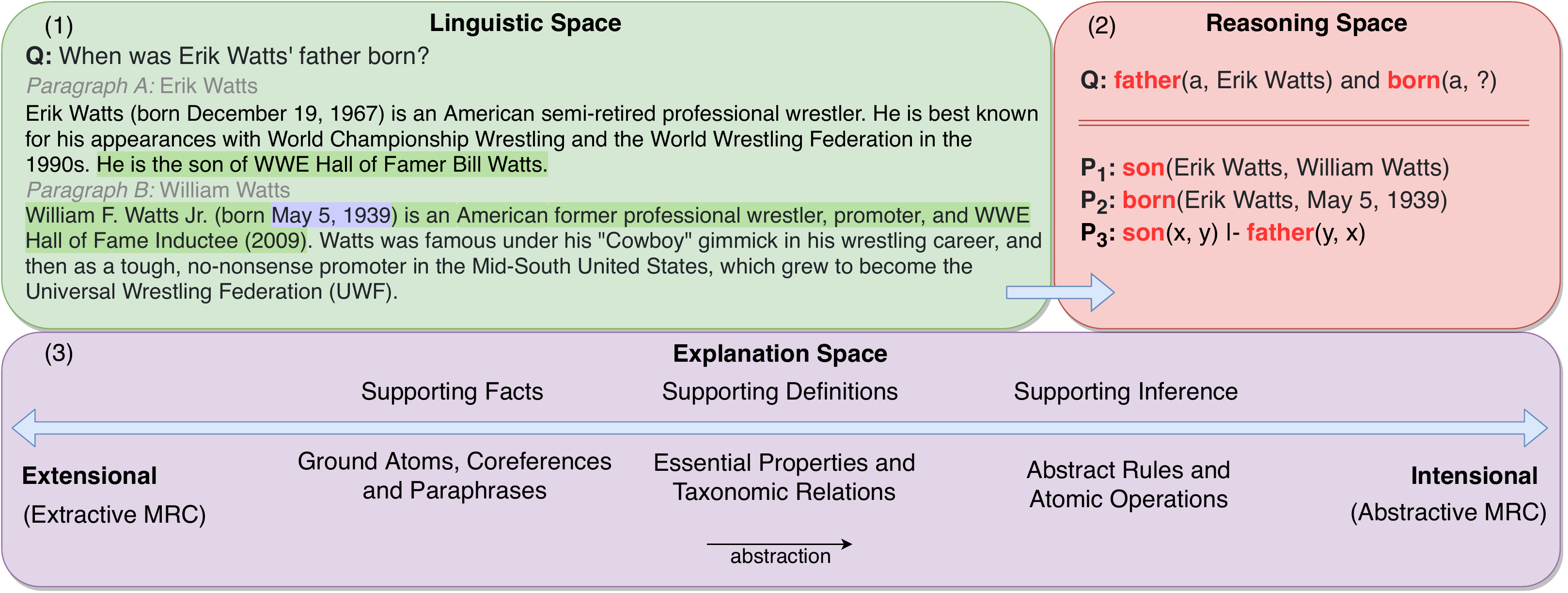}
\caption{Dimensions of explainability in Machine Reading Comprehension.}
\label{fig:approach}
\end{figure}

In the field of Explainable AI, there is no consensus, in general, on the nature of explanation \cite{miller2019explanation,biran2017explanation}. As AI embraces a variety of tasks, the resulting definition of explainability is often fragmented and dependent on the specific scenario. Here, we frame the scope of the survey by investigating the dimensions of explainability in MRC.

\paragraph{The scope of explainability.} We refer to \emph{explainability} as a specialisation of the higher level concept of \emph{interpretability}. In general, interpretability aims at developing tools to understand and investigate the behaviour of an AI system. This definition also includes tools that are external to a black-box model, as in the case of post-hoc interpretability  \cite{guidotti2018survey}. On the other hand, the goal of explainability is the design of \emph{inherently interpretable} models, capable of performing transparent inference through the generation of an \emph{explanation} for the final prediction \cite{miller2019explanation}.
In general, an explanation can be seen as an answer to a \emph{how} question formulated as follows: \emph{``How did the model arrive at the conclusion $c$ starting from the problem formulation $p$?''}. In the context of MRC, the answer to this question can be addressed by exposing the internal reasoning mechanisms linking $p$ to $c$. This goal can be achieved in two different ways: 
\begin{enumerate}
    \item \textbf{Knowledge-based explanation:} exposing part of the relevant background knowledge connecting $p$ and $c$ in terms of supporting facts and/or inference rules;  
    \item \textbf{Operational explanation:} composing a set of atomic operations through the generation of a symbolic program, whose execution leads to the final answer $c$.
\end{enumerate}

Given the scope of explainability in MRC, this survey reviews recent developments in \emph{knowledge-based} and \emph{operational explanation} (Sec. 4), emphasising the problem of \emph{explanatory relevance} for the former -- i.e. the identification of relevant information for the construction of explanations, and of \emph{question decomposition} for the latter -- i.e. casting a problem expressed in natural language into an executable program.

\paragraph{Explanation and abstraction.}

\begin{table}[t]
\small
\centering
\ra{0.9}
\begin{tabular}{@{}lp{6.5cm}p{6.5cm}@{}}
\toprule
\multirow{2}{*}{} &
\multirow{2}{*}{\textbf{Extractive MRC}} &
\multirow{2}{*}{\textbf{Abstractive MRC}}\\\\
\midrule
\textbf{Question} & When was Erik Watt's father born? & What is an example of a force producing heat?\\
\textbf{Answer} & May 5, 1939 & Two sticks getting warm when rubbed together\\
\midrule
\textbf{Explanation} & (1) He (Erik Watt) is the son of WWE Hall of Famer Bill Watts; (2) William F. Watts Jr. (born May 5, 1939) is an American former professional wrestler, promoter, and WWE Hall of Fame Inductee (2009). & (1) A stick is a kind of object; (2) To rub together means to move against; (3) Friction is a kind of force; (4) Friction occurs when two object's surfaces move against each other; (5) Friction causes the temperature of an object to increase.\\
\bottomrule
\end{tabular}
\caption{Explanations for extractive~\protect\cite{yang2018hotpotqa} and abstractive~\protect\cite{jansen2018worldtree} MRC.}
\label{tab:extractive_abstractive}
\end{table}

Depending on the nature of the MRC problem, a complete explanation can include pieces of evidence at different levels of abstraction (Fig. \ref{fig:approach}.3). Traditionally, the field has been divided into \emph{extractive} and \emph{abstractive} tasks (e.g. table \ref{tab:extractive_abstractive}). In extractive MRC, the reasoning required to solve the task is derivable from the original problem formulation. In other words, the correct decomposition of the problem provides the necessary inference steps for the answer, and the role of the explanation is to fill an information gap at the \emph{extensional level} -- i.e. identifying the correct arguments for a set of predicates, via paraphrasing and coreference resolution. As a result, explanations for extractive MRC are often expressed in the form of supporting passages retrieved from the contextual paragraphs \cite{yang2018hotpotqa}. On the other hand, abstractive MRC tasks usually require going beyond the surface form of the problem with the inclusion of high level knowledge about abstract concepts. In this case, the explanation typically leverages the use of supporting definitions, including taxonomic relations and essential properties, to perform abstraction from the original context in search of high level rules and inference patterns \cite{jansen2016s}. As the nature of the task impacts explainability, we consider the distinction between extractive and abstractive MRC throughout the survey, categorising the reviewed benchmarks and approaches according to the underlying reasoning capabilities involved in the explanations.


\section{Explanation-supporting Benchmarks}

In this section we review the benchmarks that have been designed for the development and evaluation of explainable reading comprehension models. Specifically, we classify a benchmark as \emph{explanation-supporting} if it exhibits the following properties:
\begin{enumerate}
    \item \textbf{Labelled data for training on explanations:} The benchmark includes gold explanations that can be adopted as an additional training signal for the development of explainable MRC models.
    \item \textbf{Design for quantitative explanation evaluation:} The benchmark supports the use of quantitative metrics for evaluating the explainability of MRC  systems, or it is explicitly constructed to test explanation-related inference.
\end{enumerate}

We exclude from the review all the datasets that do not comply with at least one of these requirements. For a complete overview of the existing benchmarks in MRC, the reader is referred to the following surveys:   \cite{zhang2019machine,qiu2019survey,baradaran2020survey,zhang2020machine}.
The resulting classification of the datasets with their highlighted properties is reported in Table \ref{tab:explainable_datasets}. The benchmarks are categorised according to a set of dimensions that depend on the nature of the task -- i.e. domain, format, MRC type, multi-hop inference, and the characteristics of the explanations -- i.e. explanation type, explanation level, format of the background knowledge, and explanation representation.

\begin{table}[t]
\small
\centering
\resizebox{\textwidth}{!}{
\begin{tabular}{>{\raggedright}p{3.5cm}p{12.5cm}}
    \toprule
     \textbf{Domain}    & The knowledge domain of the MRC task -- i.e. open domain (OD) , science (SCI),  or commonsense (CS).\\
     \textbf{Format} &   The task format -- i.e. span retrieval (Span), free-form (Free), multiple-choice (MC), textual entailment (TE).\\
     \textbf{MRC Type} & The reasoning capabilities involved -- i.e. Extractive (Extr.), Abstractive (Abstr.).\\
     \textbf{Multi-hop (MH) } & Whether the task requires the explicit composition of multiple facts to infer the answer.\\
     \midrule
     \textbf{Explanation Type (ET)} & The type of explanation -- i.e. knolwedge-based (KB) or operational (OP).\\
      \textbf{Explanation Level (EL)} & The abstraction level of the explanations -- i.e. Extensional (E) or Intensional (I).\\
     \textbf{Background Knowledge (BKG)} & The format of the provided background knowledge, if present, from which to extract or construct the explanations -- i.e. single paragraph (SP), multiple paragraph (MP), sentence corpus (C), table-store (TS), suit of atomic operations (AO).\\
     \textbf{Explanation Representation (ER) } & The explanation representation -- i.e. single passage (S), multiple passages (M), facts composition (FC), explanation graph (EG), generated sentence (GS), symbolic program (PR).\\
    \bottomrule\\
    \end{tabular}}
    
\small
\centering
\begin{tabular}{p{5cm}|cccc|cccc|c}
\toprule
\textbf{Dataset} &
\textbf{Domain} &
\textbf{Format} &
\textbf{Type} &
\textbf{MH} &
\textbf{ET} &
\textbf{EL} &
\textbf{BKG} &
\textbf{ER}&
\textbf{Year}\\
\midrule
\textbf{WikiQA}~\cite{yang2015wikiqa} & OD & Span & Extr. & N & KB & E & SP & S & 2015\\
\textbf{HotpotQA}~\cite{yang2018hotpotqa} & OD & Span & Extr. & Y & KB & E & MP & M & 2018\\
\textbf{MultiRC}~\cite{khashabi2018looking} & OD & MC & Abstr. & Y & KB & E & SP & M & 2018 \\
\textbf{OpenBookQA}~\cite{mihaylov2018can} & SCI & MC & Abstr. & Y & KB & I &C & FC & 2018\\
\textbf{Worldtree}~\cite{jansen2018worldtree} & SCI & MC & Abstr. & Y & KB  & I & TS & EG & 2018\\
\textbf{e-SNLI}~\cite{camburu2018snli} & CS & TE & Abstr. & N & KB  & I &  - & GS & 2018\\
\textbf{Cos-E}~\cite{rajani2019explain} & CS & MC & Abstr. & N & KB  & I & - & GS & 2019\\
\textbf{WIQA}~\cite{tandon2019wiqa} & SCI & MC & Abstr. & Y & KB  & I & SP & EG & 2019\\
\textbf{CosmosQA}~\cite{huang2019cosmos} & CS & MC & Abstr. & N & KB  & I & SP & S & 2019\\
\textbf{CoQA} \cite{reddy2019coqa} & OD & Free & Extr. & N & KB  & E & SP & S & 2019 \\
\textbf{Sen-Making}~\cite{wang2019does} & CS & MC & Abstr.& N & KB  & I & - & S & 2019\\
\textbf{ArtDataset}~\cite{bhagavatula2019abductive} & CS & MC & Abstr. &  N & KB  & I & C & S,GS & 2019\\
\textbf{QASC}~\cite{khot2020qasc} & SCI & MC & Abstr. & Y & KB  & I & C & FC & 2020\\
\textbf{Worldtree V2}~\cite{xie2020worldtree} & SCI & MC & Abstr. & Y & KB  & I & TS & EG & 2020\\
\textbf{R$^4$C}~\cite{inoue2020r4c} & OD & Span & Extr. & Y & KB  & E & MP & EG & 2020\\
\textbf{Break}~\cite{wolfson2020break} & OD & Free, Span & Abstr. & Y & OP  & I & AO & PR & 2020\\
\textbf{R$^3$}~\cite{wang2020r3} & OD & Free & Abstr. & Y & OP  & I & AO & PR & 2020\\
\bottomrule
\end{tabular}
\caption{Classification of \emph{explanation-supporting} benchmarks in MRC.}
\label{tab:explainable_datasets}
\end{table}

\paragraph{Towards abstractive and explainable MRC.}
In line with the general research trend in MRC, the development of explanation-supporting benchmarks is evolving towards the evaluation of complex reasoning, testing the models on their ability to go beyond the surface form of the text. 
Early datasets on open-domain QA have framed explanation as a \emph{sentence selection} problem \cite{yang2015wikiqa}, where the evidence necessary to infer the final answer is entirely encoded in a single supporting sentence. Subsequent work has started the transition towards more complex tasks that require the integration of multiple supporting facts.  HotpotQA \cite{yang2018hotpotqa} is one of the first \emph{multi-hop} datasets introducing a leaderboard based on a quantitative evaluation of the explanations produced by the systems\footnote{\url{https://hotpotqa.github.io/}}. The nature of HotpotQA is still closer to extractive MRC, where the supporting facts can be derived via paraphrasing from the explicit decomposition of the questions \cite{min2019multi}. 
MultiRC \cite{khashabi2018looking} combines multi-hop inference with various forms of abstract reasoning such as commonsense, causal relations, spatio-temporal and mathematical operations. The gold explanations in these benchmarks are still expressed in terms of supporting passages, leaving it implicit a consistent part of the abstract inference rules adopted to derive the answer \cite{schlegel2020framework}. Following HotpotQA and MultiRC, several benchmarks on open-domain tasks have gradually refined the supporting facts annotation, whose benefits have been demonstrated in terms of interpretability, bias, and performance \cite{dua-etal-2020-benefits,inoue2020r4c,reddy2019coqa}. Moreover, recent work have focused on complementing knowledge-based explanation with operational interpretability, introducing explicit annotation for the decomposition of multi-hop and discrete reasoning questions \cite{dua2019drop} into a sequence of atomic operations \cite{wang2020r3,wolfson2020break}.  In parallel with open-domain QA, scientific reasoning has been identified as a suitable candidate for the evaluation of explanations at a higher level of abstraction \cite{jansen2016s}. Explanations in the scientific domain naturally mention facts about underlying regularities which are hidden in the original problem formulation and that refer to knowledge about abstract conceptual categories \cite{boratko2018systematic}. The benchmarks in this domain provide gold explanations for multiple-choice science questions \cite{xie2020worldtree,jansen2018worldtree,mihaylov2018can} or related scientific tasks such as what-if questions on procedural text \cite{tandon2019wiqa} and explanation via sentences composition \cite{khot2020qasc}. Similarly to the scientific domain, a set of abstractive MRC benchmarks have been proposed for the evaluation of commonsense explanations \cite{wang2019does}. Cos-E \cite{rajani2019explain} and e-SNLI \cite{camburu2018snli} augment existing datasets for textual entailment \cite{bowman2015large} and commonsense QA \cite{talmor2019commonsenseqa} with crowd-sourced explanations, framing explainability as a natural language generation problem. Other commonsense tasks have been explicitly designed to test explanation-related inference, such as causal and abductive reasoning \cite{huang2019cosmos}. Bhagavatula et al. \shortcite{bhagavatula2019abductive} propose the tasks of Abductive Natural Language Inference ($\alpha$NLI) and Abductive Natural Language Generation ($\alpha$NLG), where MRC models are required to select or generate the hypothesis that best explains a set of observations.

\paragraph{Multi-hop reasoning and explanation.}
The ability to construct explanations in MRC is typically associated with multi-hop reasoning. However, the nature and the structure of the inference can differ greatly according to the specific task. In extractive MRC \cite{yang2018hotpotqa,welbl2018constructing}, multi-hop reasoning often consists in the identification of bridge entities, or in the extraction and comparison of information encoded in different passages. On the other hand, \cite{jansen2018worldtree} observe that complete explanations for science questions require an average of 6 facts classified in three main explanatory roles: \emph{grounding facts} and \emph{lexical glues} have the function of connecting the specific concepts in the question with abstract conceptual categories, while \emph{central facts} refer to high-level explanatory knowledge. Similarly, OpenbookQA \cite{mihaylov2018can} provides annotations for the core explanatory sentences, which can only be inferred by performing multi-hop reasoning through the integration of external commonsense knowledge. In general, the number of hops needed to construct the explanations is correlated with \emph{semantic drift} -- i.e. the tendency of composing spurious inference chains that lead to wrong conclusions \cite{khashabi2019capabilities,fried2015higher}.  Recent explanation-supporting benchmarks attempt to limit this phenomenon by providing additional signals to learn abstract composition schemes, via the explicit annotation of valid inference chains \cite{khot2020qasc} or the identification of common explanatory patterns \cite{xie2020worldtree}.

\section{Explainable MRC Architectures}

\begin{figure}[t]%
\centering
\subfloat[Explainable Abstractive MRC Approaches]{{\includegraphics[width=0.46\textwidth]{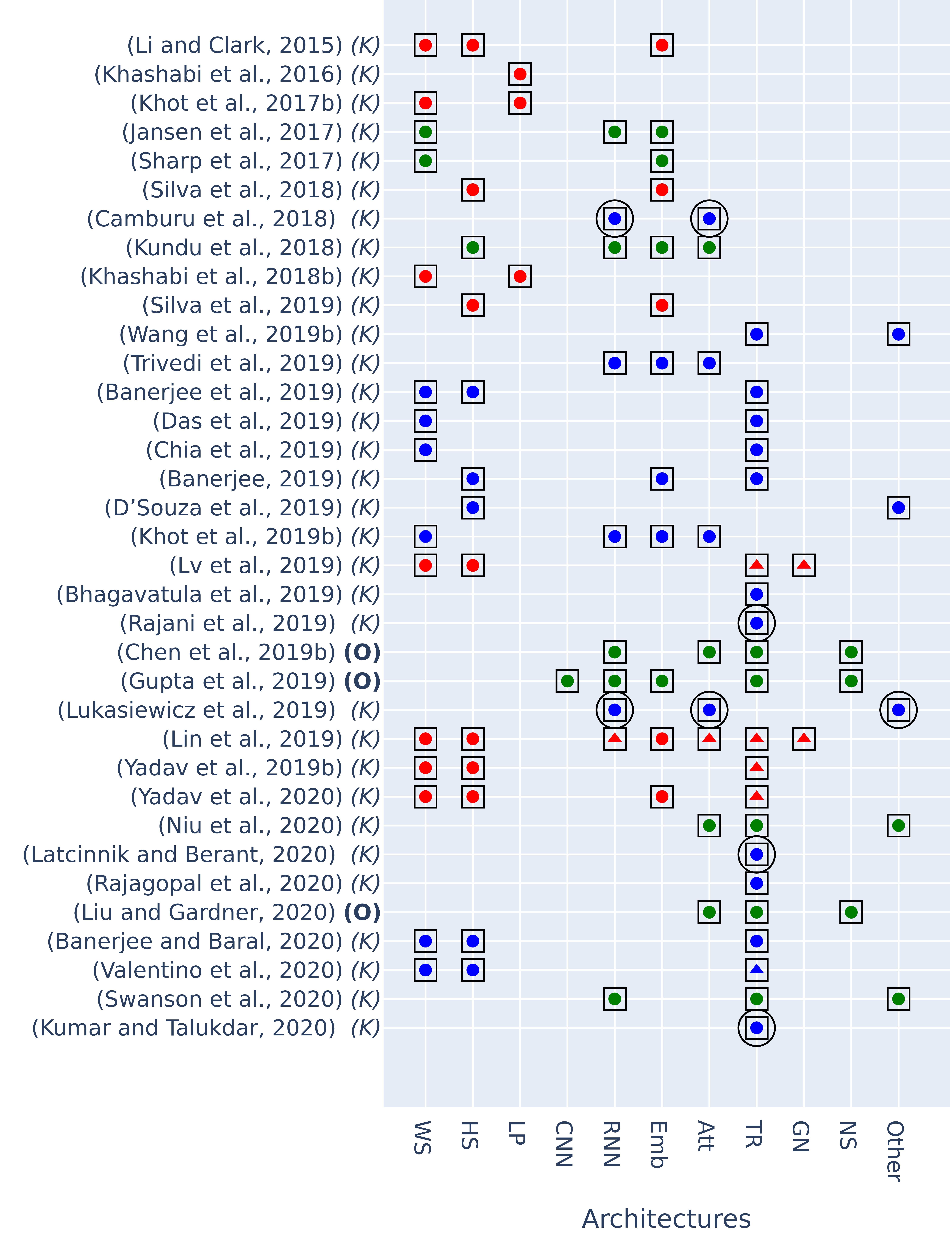} }}%
\qquad
\subfloat[Explainable Extractive MRC Approaches]{{\includegraphics[width=0.46\textwidth]{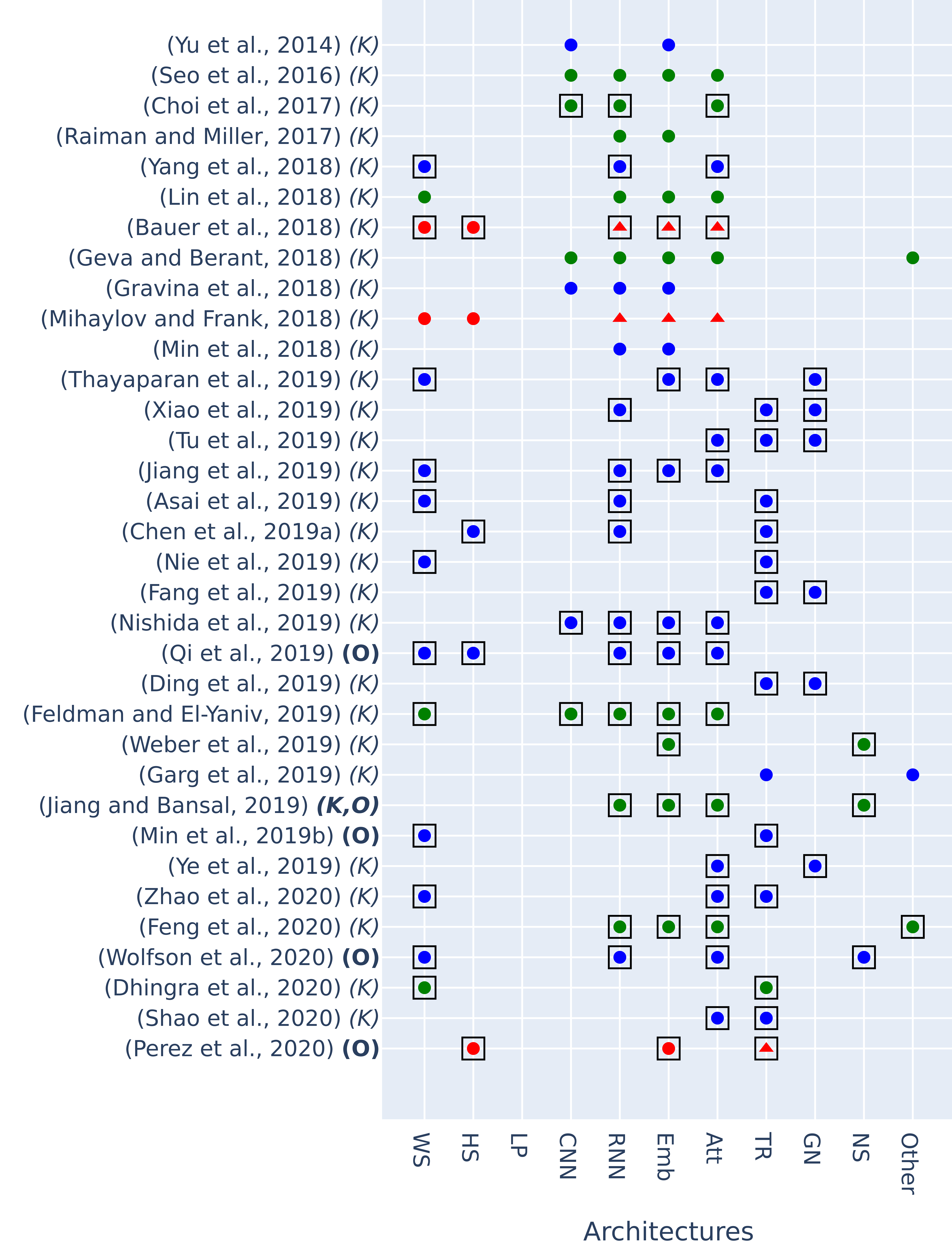} }}%
\caption{\small Explainable Machine Reading Comprehension (MRC) approaches. \textbf{Operational Explanations}: (\textbf{O}), \textbf{Knowledge-based Explanations}: (\textit{K}), \textbf{Operational and Knowledge-based Explanations}: (\textbf{\textit{K,O}})
\textbf{Learning}: Unsupervised (\tikzcircle{3pt}), Distantly Supervised (\tikzcircle{3pt,OliveGreen}), Strongly Supervised (\tikzcircle{3pt,Blue}).
\textbf{Generated Output}: (\tikzcircle[black,fill=white]{4pt}). \textbf{Multi Hop}: ($\fbox{$\phantom{5}$}$). \textbf{Answer Selection Module:} ($\bigtriangleup$). \textbf{Architectures}: \textsc{Weighting Schemes} (\underline{WS}):  Document and query weighting schemes consist of information retrieval systems that use any form of vector space scoring system, \textsc{Heuristics} (\underline{HS}): Hand-coded heuristics and scoring functions, \textsc{Linear Programming} (\underline{LP}), \textsc{Convolutional Neural Network} (\underline{CNN}), \textsc{Recurrent Neural Networks} (\underline{RNN}), \textsc{Pre-Trained Embeddings} (\underline{Emb}), \textsc{Attention Network} (\underline{Att}), \textsc{Transformers} (\underline{TR}), \textsc{Graph Neural Networks} (\underline{GN}), \textsc{Neuro-Symbolic} (\underline{NS}) and \textsc{Others}.
}%
\label{fig:approaches}
\end{figure}

This section describes the major architectural trends for Explainable MRC (X-MRC). The approaches are broadly classified according to the nature of the MRC task they are applied to -- i.e. extractive or abstractive. In order to elicit architectural trends, we further categorise the approaches as described in Table~\ref{tab:approach_category}.
\begin{table}[t]
    \centering
    \small
    \resizebox{\textwidth}{!}{
    \begin{tabular}{>{\raggedright}p{2.5cm}p{13cm}}
    \toprule
     \textbf{Explanation Type} & (1) Knowledge-based explanation; (2) Operational-based explanation \\
     \midrule
     \textbf{Learning method} & (1) Unsupervised (US): Does not require any annotated data; (2) Strongly Supervised (SS): Requires gold explanations for training or inference; (3) Distantly Supervised (DS): Treats explanation as a latent variable training only on problem-solution pairs.\\
     \midrule
     \textbf{Generated Output} & Denotes whether the explanation is generated or composed from facts retrieved from the background knowledge.\\
     \midrule
     \textbf{Multi-Hop} & Denotes whether the approach is designed for multi-hop reasoning \\
    \bottomrule
    \end{tabular}}
    \caption{Categories adopted for the classification of Explainable MRC approaches.}
    \label{tab:approach_category}
\end{table}
Figure~\ref{fig:approaches} illustrates the resulting classification when 
considering the underlying architectural components.
If an approach employs distinct modules for explanation generation and answer prediction, the latter is marked as $\bigtriangleup$. For these instances, we only consider the categorization for the explanation extraction module.

Admittedly, the boundaries of these categories can be quite fuzzy. For instance, pre-trained embeddings such as ELMo~\cite{Peters:2018} are composed of recurrent neural networks, and transformers are composed of attention networks. In cases like these, we only consider the larger component that subsumes the smaller one. If approaches employ both architectures, but as different functional modules, we plot them separately. 

In general, we observe an overall shift towards supervised methods over the years for both abstractive and extractive MRC. 
We posit that the advent of explanation-supporting datasets has facilitated the adoption of complex supervised neural architectures.
Moreover, as shown in the classification, the majority of the approaches are designed for knowledge-based explanation. We attribute this phenomenon to the absence of large-scale datasets for operational interpretability until 2020. However, we note a recent uptake of distantly supervised approaches. We believe that further progress can be made with the introduction of novel datasets supporting symbolic question decomposition such as Break \cite{wolfson2020break} and R$^3$ \cite{wang2020r3} (See Sec.~\ref{tab:explainable_datasets}).



\subsection{Modeling Explanatory Relevance for Knowledge-based Explanations}

This section reviews the main approaches adopted for modeling \emph{explanatory relevance}, namely the problem of identifying the relevant information for the construction of \emph{knowledge-based explanations}. 
We group the models into three main categories: \textit{Explicit}, \textit{Latent}, and \textit{Hybrid}. 

\subsubsection{Explicit Models}


Explicit models typically adopt heuristics and hand-crafted constraints to encode high level hypotheses of explanatory relevance. The major architectural patterns are listed below:

\paragraph{Linear Programming (LP):}
Linear programming has been used for modeling semantic and structural constrains in an unsupervised fashion.
Early LP systems, such as TableILP~\cite{khashabi2016question}, formulate the construction of explanations as an optimal sub-graph selection problem over a set of semi-structured tables. Subsequent approaches~\cite{khot2017answering,khashabi2018question} 
have proposed methods to reason over textual coprora via semantic abstraction, leveraging semi-structured representations automatically extracted through Semantic Role Labeling, OpenIE, and Named Entity Recognition. 
Approaches based on LP have been effectively applied for multiple-choice science questions, when no gold explanation is available for strong supervision. 

\paragraph{Weighting schemes with heuristics:} The integration of heuristics and weighing schemes have been demonstrated to be effective for the implementation of lightweight methods that are inherently scalable to large corpora and knowledge bases. 
In the open-domain, approaches based on lemma overlaps and weighted triplet scoring function have been proposed~\cite{mihaylov2018knowledgeable}, along with path-based heuristics implemented with the auxiliary use of external knowledge bases ~\cite{bauer2018commonsense}. 
Similarly, path-based heuristics have been adopted for commonsense tasks, where Lv et al.~\shortcite{lv2019graph} propose a path extraction technique based on question coverage. For scientific and multi-hop MRC, Yadav et al.~\shortcite{yadav2019quick} propose ROCC, an unsupervised method to retrieve multi-hop explanations that maximise relevance and coverage while minimising overlaps between intermediate hops. Valentino et al.~\shortcite{valentino2020unification,valentino2020explainable} present an explanation reconstruction framework for multiple-choice science questions based on the notion of unification in science. The unification-based framework models explanatory relevance using two scoring functions: a relevance score representing lexical similarity, and a unification score denoting the explanatory power of fact, depending on its frequency in explanations for similar cases. 

\paragraph{Pre-trained embeddings with heuristics:} Pre-trained embeddings have the advantage of capturing semantic similarity, going beyond the lexical overlaps limitation imposed by the use of weighting schemes. This property has been shown to be useful for multi-hop and abstractive tasks, where approaches based on pre-trained word embeddings, such as GloVe \cite{pennington-etal-2014-glove}, have been adopted to perform semantic alignment between question, answer and justification sentences \cite{yadav2020unsupervised}. 
Silva et al.~\shortcite{silva2018recognizing,silva2019exploring} employ word embeddings and semantic similarity scores to perform selective reasoning on commonsense knowledge graphs and construct explanations for textual entailment. Similarly, knowledge graph embeddings, such as TransE~\cite{wang2014knowledge}, have been adopted for extracting reasoning paths for commonsense QA~\cite{lin2019kagnet}.


\subsubsection{Latent Models}
\label{sec:latent_model}

Latent models 
learn the notion of explanatory relevance implicitly through the use of machine learning techniques such as neural embeddings and neural language models. The architectural clusters adopting latent modeling are classified as follows:

\paragraph{Neural models for sentence selection:} This category refers to a set of neural approaches proposed for the \emph{answer sentence selection} problem. 
These approaches typically adopt deep 
learning architectures, such as RNN, CNN and Attention networks via strong or distant supervision.  Strongly supervised approaches~\cite{yu2014deep,min2018efficient,gravina2018cross,garg2019tanda} are trained on gold supporting sentences. In contrast, distantly supervised techniques indirectly learn to extract the supporting sentence by training on the final answer. Attention mechanisms have been frequently used for distant supervision~\cite{seo2016bidirectional} to highlight the attended explanation sentence in the contextual passage. Other distantly supervised approaches model the sentence selection problem through the use of latent variables~\cite{raiman2017globally}.

\paragraph{Transformers for multi-hop reasoning:} 
Transformers-based architectures \cite{vaswani2017attention} have been successfully applied to learn explanatory relevance in both extractive and abstractive MRC tasks. Banerjee~\shortcite{banerjee2019asu} and Chia et al.~\shortcite{chia2019red} adopt a BERT model \cite{devlin2018bert} to learn to rank explanatory facts in the scientific domain. Shao et al.~\shortcite{shao2020graph} employ transformers with self-attention on multi-hop QA datasets \cite{yang2018hotpotqa}, demonstrating that the attention layers implicitly capture high-level relations in the text. The Quartet model \cite{rajagopal2020if} has been adopted for reasoning on procedural text and 
producing structured explanations based on qualitative effects and interactions between concepts.
In the distant supervison setting, Niu et al.~\shortcite{niu2020self} address the problem of lack of gold explanations by training a self-supervised evidence extractor with auto-generated labels in an iterative process. Banerjee and Bara \shortcite{banerjee2020knowledge} propose a semantic ranking model based on BERT for QASC \cite{khot2020qasc} and OpenBookQA \cite{mihaylov2018can}. Transformers have shown improved performance on downstream answer prediction tasks when applied in combination with explanations constructed through explicit models~\cite{yadav2019quick,yadav2020unsupervised,valentino2020unification}.

\paragraph{Attention networks for multi-hop reasoning:} Similar to transformer-based approaches, attention networks have also been employed to extract relevant explanatory facts. However, attention networks are 
usually applied in combination with other neural modules. For HotpotQA, Yang et al.~\shortcite{yang2018hotpotqa}  propose a model trained in a multi-task setting on both gold explanations and answers, composed of recurrent neural networks and attention layers. 
Nishida et al.~\shortcite{nishida2019answering} introduce a similarly structured model with a query-focused extractor designed to elicit explanations. The distantly supervised MUPPET model~\cite{feldman2019multi} captures the relevance between question and supporting facts through bi-directional attention on sentence vectors encoded using pre-trained embedding, CNN, and RNN. In the scientific domain, Trivedi et al.~\shortcite{trivedi2019repurposing} repurpose existing textual entailment datasets to learn the supporting facts relevance for multi-hop QA. Khot et al.~\shortcite{khot2019s} propose a knowledge gap guided framework to construct explanations for OpenBookQA.  

\paragraph{Language generation models:} 
Recent developments in language modeling along with the creation of explanation-supporting benchmarks, such as e-SNLI \cite{camburu2018snli} and Cos-E \cite{rajani2019explain}, have opened up the possibility to automatically generate semantically plausible and coherent explanation sentences. 
Language models, such as GPT-2~\cite{radford2019language}, have been adopted for producing commonsense explanations, whose application has demonstrated benefits in terms of accuracy and zero-shot generalisation \cite{rajani2019explain,latcinnik2020explaining}. Kumar and Talukdar~\shortcite{kumar2020nile} present a similar approach for natural language inference, generating explanations for entailment, neutral and contradiction labels. e-SNLI~\cite{camburu2018snli} present a baseline based on a Bi-LSTM encoder-decoder with attention. Lukasiewicz et al.~\shortcite{lukasiewiczmake} enhance this baseline by proposing an adversarial framework to generate more consistent and plausible explanations. 

\subsubsection{Hybrid Models}

Hybrid models adopt heuristics and hand-crafted constraints as a pre-processing step to impose an explicit inductive bias for explanatory relevance. The major architectural patterns are listed below:

\paragraph{Graph Networks:}

The relational inductive bias encoded in Graph Networks \cite{battaglia2018relational} provides a viable support for reasoning and learning over structured representations. This characteristic has been identified as particularly suitable for supporting facts selection in multi-hop MRC tasks. A set of graph-based architectures have been proposed for multi-hop reasoning in HotpotQA \cite{yang2018hotpotqa}. Ye et al.~\shortcite{ye2019multi} build a graph using sentence vectors as nodes 
and edges connecting sentences that share the same named entities. Similarly, Tu et al.~\shortcite{tu2019select} construct a graph connecting sentences that are part of the same document, share noun-phrases, and have named entities or noun phrases in common with the question. 
Thayaparan et al. \shortcite{thayaparan2019identifying} propose a graph structure including both documents and sentences as nodes. The graph connects documents that mention the same named entities. To improve scalability, the Dynamically Fused Graph Network (DFGN)~\cite{xiao2019dynamically} adopts a dynamic construction of the graph, starting from the entities in the question and gradually selecting the supporting facts. Similarly, Ding et al.~\shortcite{ding2019cognitive} implement a dynamic graph exploration inspired by the dual-process theory~\cite{evans2003two,sloman1996empirical,evans1984heuristic}. 
the Hierarchical Graph Network \cite{fang2019hierarchical} leverages a hierarchical graph representation of the background knowledge (i.e. question, paragraphs, sentences, and entities). In parallel with extractive MRC tasks, Graph Networks are applied for answer selection on commonsense reasoning, where a subset of approaches have started exploring the use of explanation graphs extracted from external knowledge bases through path-based heuristics \cite{lv2019graph,lin2019kagnet}.

\paragraph{Explicit inference chains for multi-hop reasoning:} A subset of approaches has introduced end-to-end frameworks explicitly designed to emulate the step-by-step reasoning process involved in multi-hop MRC~\cite{kundu2018exploiting,chen2019multi,jiang2019explore}.  The baseline approach proposed for Abductive Natural Language Inference~\cite{bhagavatula2019abductive}  builds chains composed of hypotheses and observations, and encode them using transformers to identify the most plausible explanatory hypothesis.  Similarly, Das et al.~\shortcite{das2019chains} embed the reasoning chains retrieved via TF-IDF and lexical overlaps using a BERT model to identify plausible explanatory patterns for multiple-choice science questions. In the open domain, Asai et al.~\shortcite{asai2019learning} build a graph structure using entities and hyperlinks and adopt recurrent neural networks to retrieve relevant documents sequentially. Nie et al.~\shortcite{nie2019revealing} introduce a step-by-step reasoning process that first retrieves the relevant paragraph, then the supporting sentence, and finally, the answer.  Dhingra et al.~\shortcite{dhingra2020differentiable} propose an end-to-end differentiable model that uses Maximum Inner Product Search (MIPS)~\cite{johnson2019billion} to query a virtual knowledge-base and extract a set of reasoning chains. Feng et al~\shortcite{feng2020learning} propose a cooperative game approach to select the most relevant explanatory chains from a large set of candidates. In contrast to neural-based methods, Weber et al.~\shortcite{weber2019nlprolog} propose a neuro-symbolic approach for multi-hop reasoning that extends the unification algorithm in Prolog with pre-trained sentence embeddings.

\subsection{Operational Explanation}

Operational explanations aim at providing  interpretability by exposing the set of operations adopted to arrive at the final answer. This section reviews the main architectural patterns for operational interpretability that focus on the problem of casting a question into an executable program.

\paragraph{Neuro-Symbolic models:} Neuro-symbolic approaches combine neural models with symbolic programs.
Liu and Gardner~\shortcite{liu2020multi} propose a multi-step inference model with three primary operations: Select, Chain, and Predict. The Select operation retrieves the relevant knowledge; the Chain operation composes the background knowledge together; the Predict operation select the final answer. Jiang and Bansel.~\shortcite{jiang2019self} propose the adoption of Neural Module Networks~\cite{andreas2016neural} for multi-hop QA by designing four atomic neural modules (Find, Relocate, Compare, NoOp) that allow for both operational explanation and supporting facts selection. Similarly, Gupta et al.~\shortcite{gupta2019neural} adopt Neural Module Networks to perform discrete reasoning on DROP \cite{dua2019drop}. In contrast, Chen et al.~\shortcite{chen2019neural} propose an architecture based on LSTM, attention modules, and transformers to generate compositional programs. While most of the neuro-symbolic approaches are distantly supervised, the recent introduction of question decomposition datasets~\cite{wolfson2020break} allows for a direct supervision of symbolic program generation \cite{subramanian2020obtaining}.

\paragraph{Multi-hop question decomposition:} The approaches in this category aim at breaking multi-hop questions into single-hop queries that are simpler to solve. The decomposition allows for the application of divide-et-impera methods where the solutions for the single-hop queries are computed individually and subsequently merged to derive the final answer. Perez et al.~\shortcite{perez2020unsupervised} propose an unsupervised decomposition method for the HotpotQA dataset. Min et al.~\shortcite{min2019multi} frame question decomposition as a span prediction problem adopting supervised learning with a small set of annotated data. Qi et al.~\shortcite{qi2019answering} propose GOLDEN Retriever, a scalable method to generate search queries for multi-hop QA, enabling the application of off-the-shelf information retrieval systems for the selection of supporting facts.

\section{Evaluation}
\label{sec:evaluation}
The development of explanation-supporting benchmarks has allowed for a quantitative evaluation of the explainability in MRC. In open-domain settings, Exact Matching (EM) and F1 score are often employed for evaluating the supporting facts \cite{yang2018hotpotqa}, while explanations for multiple-choice science questions have been evaluated using ranking-based metrics such as Mean Average Precision (MAP) \cite{xie2020worldtree,jansen-ustalov-2019-textgraphs}. In contexts where the explanations are produced by language models, natural language generation metrics have been adopted, such as BLEU score and perplexity \cite{papineni2002bleu,rajani2019explain}. Human evaluation still plays an important role, especially for distantly supervised approaches applied on benchmarks that do not provide labelled explanations. 


\subsection{Silver Explanations}

Since annotating explanations are expensive and not readily available, approaches automatically curate \textit{silver} explanations for training. For single-hop Extractive MRC, both \cite{raiman2017globally} and \cite{min2018efficient} use oracle sentence (the sentence containing the answer span) as the descriptive explanation. Similarly, for multi-hop Extractive approaches~\cite{chen2019multi,wang2019evidence} extract explanations by building a path connecting question to the oracle sentence, by linking multiple sentences using inter-sentence knowledge representation. Since there might be multiple paths connecting question and answer, to determine the best path, ~\cite{chen2019multi} uses the shortest path with the highest lexical overlap and ~\cite{wang2019evidence} employs Integer Linear Programming (ILP) with hand-coded heuristics.

\paragraph{Evaluating multi-hop reasoning} Evaluating explainability through multi-hop reasoning presents still several challenges \cite{chen2019understanding,wang2019multi}. Recent works have demonstrated that some of the questions in multi-hop QA datasets do not require multi-hop reasoning or can be answered by exploiting statistical shortcuts in the data \cite{min2019compositional,chen2019understanding,jiang2019avoiding}. In parallel, other works have shown that a consistent part of the expected reasoning capabilities for a proper evaluation of reading comprehension are missing in several benchmarks \cite{schlegel2020framework,kaushik2018much}. A set of possible solutions have been proposed to overcome some of the reported issues, including the creation of evaluation frameworks for the gold standards \cite{schlegel2020framework}, the development of novel metrics for multi-hop reasoning \cite{trivedi2020measuring}, and the adoption of adversarial training techniques \cite{jiang2019avoiding}. A related research problem concerns the faithfulness of the explanations. Subramanian et al. \shortcite{subramanian2020obtaining} observe that some of the modules in compositional neural networks \cite{andreas2016neural}, particularly suited for operational interpretability, do not perform their intended behaviour. To improve faithfulness the authors suggest novel architectural design choices and propose the use of auxiliary supervision. 

\section{Conclusion and Open Research Questions}


This survey has proposed a systematic categorisation of benchmarks and approaches for explainability in MRC. Lastly, we outline a set of open research questions for future work:  

\begin{enumerate}
    \item \textbf{Contrastive  Explanations}: while contrastive and conterfactual explanations are becoming central in Explainable AI \cite{miller2019explanation}, this type of explanations is still under-explored for MRC. We believe that contrastive explanations can lead to the development of novel reasoning paradigms, especially in the context of multiple-choice science and commonsense QA
    \item \textbf{Benchmark Design:} to advance research in explainability it is fundamental to develop reliable methods for explanation evaluation, overcoming the issues presented in Section \ref{sec:evaluation}, and identify techniques for scaling up the annotation of gold explanations \cite{inoue2020r4c}
    \item \textbf{Knowledge Representation:} the combination of explicit and latent representations have been useful for explainability in multi-hop, extractive MRC. An open research question is understanding whether a similar paradigm can be beneficial for abstractive tasks to limit the phenomenon of semantic drift observed in many hops reasoning
    \item \textbf{Supervised Program Generation:} large-scale benchmarks for operational explanations have been released just recently \cite{wolfson2020break,wang2020r3}. We believe that these corpora open up the possibility to explore strongly supervised methods to improve accuracy and faithfulness in compositional neural networks and symbolic program generation.
\end{enumerate}





\bibliographystyle{acl}
\bibliography{coling2020,approach}

\end{document}